\documentclass{article}

\usepackage{amsmath}
\usepackage{booktabs}
\usepackage{multirow}
\usepackage{makecell}
\usepackage{xcolor}
\usepackage{graphicx}
\usepackage{subcaption}
\usepackage[preprint]{neurips_2026}

\usepackage[utf8]{inputenc} 
\usepackage[T1]{fontenc}    
\usepackage{hyperref}       
\usepackage{url}            
\usepackage{booktabs}       
\usepackage{amsfonts}       
\usepackage{nicefrac}       
\usepackage{microtype}      
\usepackage{xcolor}         

\title{Mix-QVLA: Task-Evidence-Aware Mixed-Precision Quantization of Vision-Language-Action Models}

\author{Navin Ranjan \hspace{0.75cm} Andreas Savakis\\
Rochester Institute of Technology\\
Rochester, New York 14623, USA\\
{\tt\small nr4325@rit.edu \hspace{0.5cm} andreas.savakis@rit.edu}}

\begin{document}

\maketitle

\begin{abstract}
Vision-language-action (VLA) models unify perception, language reasoning, and robot control within a single policy, but their memory and compute requirements limit deployment on resource-constrained robotic platforms. Low-bit mixed-precision post-training quantization (PTQ) offers a practical path for compression. However, existing VLA quantization methods primarily estimate layer sensitivity from final action deviation, leaving task-relevant evidence preservation across internal policy stages largely unexamined. As a result, a quantized model may produce a similar motion command while disrupting the evidence structure that supports the full-precision policy decision. We propose \textbf{Mix-QVLA}, a task-evidence-aware mixed-precision PTQ framework for VLA models. Mix-QVLA anchors each quantized variant to the full-precision action-token reference decision and evaluates whether quantization preserves task-relevant evidence across key VLA functional boundaries. It computes normalized gradient-weighted task-evidence maps from boundary activations and compares full-precision and quantized maps using evidence-mass and attribution-distribution distortion, capturing changes in both the strength and allocation of decision-supporting evidence. A soft-bottleneck objective aggregates boundary-level degradation into layer-wise sensitivity scores. Mix-QVLA further models sensitivity throughout task execution, capturing phase-dependent shifts in layer importance rather than assuming a fixed sensitivity profile. The resulting evidence- and time-aware scores guide mixed-precision bit allocation under model-size and BitOps budgets. Extensive evaluations on OpenVLA-style policies show that Mix-QVLA improves the accuracy--efficiency trade-off of low-bit VLA deployment. On LIBERO, Mix-QVLA reduces OpenVLA-OFT memory from 15.4 GB to 4.1 GB, retains 96.3\% average success compared with 97.1\% for the BF16 model, and achieves a 1.52× inference speedup.
\end{abstract}

\section{Introduction}
Vision-language-action (VLA) models 
provide a unified interface for embodied multimodal intelligence, mapping visual observations and natural-language instructions directly to executable robot actions. By integrating perception, language-conditioned reasoning, and control within a single policy, recent systems such as OpenVLA~\cite{kim2024openvla}, OpenVLA-OFT~\cite{kim2025fine}, $\pi_{0.5}$~\cite{intelligence2025pi_} and GR00T N1.5~\cite{bjorck2025gr00t} highlight the potential of VLA models for language-conditioned manipulation and general-purpose robot policies. However, this progress comes with increasing computational demands, making efficient deployment a central challenge. For example, a 7B-parameter OpenVLA model in half precision is already on the order of 14 GB, placing substantial pressure on model scale, inference latency, and hardware accessibility for resource-constrained robotic platforms. These constraints have motivated recent work on VLA efficiency, including pruning and quantization~\cite{yang2025efficientvla,xu2026qvla,zheng2026dyq,zhang2026quantvla,xu2026ptq}, as well as efficient VLA system design~\cite{wen2025tinyvla,song2025accelerating,zhang2026mole,shukor2025smolvla}.

Post-training quantization (PTQ) is an attractive direction for VLA compression because it can reduce model footprint and inference cost without retraining the full model. Existing PTQ methods for large language and vision-language models have made substantial progress by preserving weight reconstruction, controlling activation outliers, or identifying salient channels~\cite{frantar2022gptq,xiao2023smoothquant,lin2024awq}. However, VLA quantization introduces a distinct sensitivity problem. Unlike text generation or image classification, a VLA output is executed as a robot action and can influence future observations through closed-loop interaction. As a result, quantization errors can propagate through visual grounding, language-conditioned reasoning, and action-token prediction, ultimately affecting the stability of the resulting control trajectory. This makes it difficult to directly transfer generic LLM or VLM quantization criteria to embodied policies.

Recent VLA-aware quantization work has recognized that embodied policies require criteria beyond generic LLM or VLM reconstruction objectives. QVLA~\cite{xu2026qvla}, for example, introduces an action-centric quantization framework that uses final action variation as the primary signal for estimating sensitivity. Other recent methods improve VLA quantization through temporal precision adaptation or scale-calibrated PTQ mechanisms~\cite{zheng2026dyq,zhang2026quantvla}. These works highlight the need to design quantization methods around the embodied nature of VLA policies. However, existing criteria still provide an incomplete view of quantization sensitivity: action deviation captures the endpoint of the policy computation, kinematic signals describe runtime execution dynamics, and scale calibration stabilizes low-bit inference, but none directly measures whether quantization preserves the internal evidence pathway supporting the full-precision policy decision. A small action deviation does not guarantee that the quantized model preserves this decision pathway, while a large deviation does not reveal where the VLA computation failed.

This motivates \textbf{Mix-QVLA}, a task-evidence-aware mixed-precision PTQ framework for VLA models. Mix-QVLA evaluates quantization sensitivity by measuring whether the task evidence supporting the full-precision action decision is preserved across major functional boundaries of the VLA pipeline, including visual encoding, vision-language projection, multimodal policy reasoning, and pre-action decision formation. Rather than relying only on final action deviation, Mix-QVLA anchors evidence to the full-precision decision and measures how quantization disrupts the internal support for that decision before the final action is produced. It compares full-precision and quantized evidence through complementary evidence-preservation measures and aggregates boundary-level degradation into layer-wise sensitivity scores. Because VLA policies unfold over trajectories, Mix-QVLA further treats sensitivity as time-dependent and accumulates evidence degradation across control steps. The resulting scores guide constrained mixed-precision bit allocation, preserving higher precision for layers most critical to the full-precision decision pathway under model-size and BitOps constraints. Our main contributions are as follows:

\begin{itemize}
\item We propose \textbf{Mix-QVLA}, a task-evidence-aware mixed-precision PTQ framework for VLA models. It estimates layer sensitivity from evidence preservation across functional boundaries and trajectory timesteps, then uses these scores for budget-constrained bit allocation.
\item We introduce boundary-level task-evidence analysis to evaluate how quantization disrupts the internal VLA decision pathway. By measuring evidence preservation across functional stages, it provides layer-wise sensitivity estimates for mixed-precision quantization and diagnostic insight into how VLA components support the full-precision policy decision.
\item We introduce a time-aware sensitivity analysis that captures phase-dependent layer importance across trajectory steps, distinguishing layers that are consistently fragile from those that become sensitive only at specific stages of execution.
\item We validate Mix-QVLA on OpenVLA-style policies, showing improved sensitivity estimation and stronger success--efficiency trade-offs than action-only quantization criteria.
\end{itemize}

\section{Related Work}
\paragraph{Vision-Language-Action Models.}
Recent VLA models formulate robotic control as vision-language-conditioned action prediction. Existing systems are commonly distinguished by their action decoding strategy. Token-based models, such as RT-2 (~\cite{zitkovich2023rt}), OpenVLA (~\cite{kim2024openvla}), and UniVLA (~\cite{bu2025univla}), discretize continuous robot actions and cast control as sequence generation. RT-2 transfers web-scale vision-language knowledge to robotic actions, OpenVLA scales this direction with an open 7B autoregressive VLA policy, and UniVLA studies unified modeling across vision, language, and action. In contrast, generative-action models, such as Octo (~\citep{team2024octo}), RDT-1B (~\citep{liu2024rdt}), and $\pi_0$ (~\citep{black2024pi_0}), predict continuous actions using diffusion- or flow-based decoders. These models improve action expressiveness and temporal control. However, these systems have substantial computational footprints. Recent methods reduce VLA inference cost through compact architectures, pruning, caching, or early-exit decoding. TinyVLA (~\citep{wen2025tinyvla}) designs a smaller VLA backbone for efficient manipulation. EfficientVLA (~\citep{yang2025efficientvla}) combines language-layer pruning, visual token selection, and diffusion-head caching to reduce redundant computation. CEED-VLA (~\citep{song2025ceed}) accelerates action generation using consistency-based early-exit decoding. These methods improve inference efficiency, but they mainly focus on reducing runtime computation rather than analyzing how compression affects the action decisions.

\paragraph{Quantization.}
Post-training quantization (PTQ) compresses pretrained models by calibrating quantization parameters on a small dataset, avoiding expensive retraining. Recent transformer quantization methods, such as SmoothQuant and OmniQuant, address activation outliers, scale imbalance, and low-bit calibration challenges~\citep{xiao2023smoothquant,shao2023omniquant}. However, these methods are mainly developed for LLMs or VLMs, where the objective is typically to preserve text generation, classification, or multimodal prediction quality. They do not directly account for the closed-loop action behavior of VLA policies, where small quantization-induced changes can alter grounding, reasoning, or control over time. Recent VLA-specific methods address this gap from different perspectives. EaqVLA proposes encoding-aligned quantization for token alignment (~\citep{jiang2025eaqvla}); QuantVLA introduces scale-calibrated PTQ with selective quantization, attention temperature matching, and output-head balancing (~\citep{zhang2026quantvla}); QVLA estimates sensitivity from final action deviation (~\citep{xu2026qvla}); and DyQ-VLA adjusts precision over time using kinematic proxies (~\citep{zheng2026dyq}). Although effective, these methods mainly rely on action-centric, scale-centric, or proxy-based sensitivity signals. Such signals may miss cases where quantization preserves a similar final action but disrupts the internal evidence supporting the full-precision decision, causing incorrect task grounding even when the motion remains smooth. Our work addresses this limitation by measuring task-evidence preservation across visual grounding, language-conditioned reasoning, and pre-action decision formation.

\section{Method}

\textbf{Vision-language-action models.}
A vision-language-action model parameterizes the control policy of an embodied agent by mapping multimodal inputs to robot actions. 
At environment timestep $\tau$, the VLA input consists of an RGB visual observation $V_{\tau}$, a task instruction $P$, and a robot state $x_{\tau}$. We denote the multimodal input as $z_{\tau}=(V_{\tau},x_{\tau},P)$. The policy parameterized by $\theta$ predicts a tokenized robot action $y_{\tau}=(y_{\tau,1},\ldots,y_{\tau,K})$ according to $\pi_{\theta}(y_{\tau}\mid z_{\tau})$, where $K$ denotes the number of action tokens. The policy assigns likelihood to this sequence through the autoregressive factorization
\begin{equation}
    \pi_{\theta}(y_{\tau}\mid z_{\tau})
    =
    \prod_{k=1}^{K}
    p_{\theta}
    \left(
        y_{\tau,k}
        \mid
        y_{\tau,<k}; z_{\tau}
    \right),
\end{equation}
where $y_{\tau,<k}=(y_{\tau,1},\ldots,y_{\tau,k-1})$ is the action-token prefix. The predicted token sequence is finally mapped to a continuous robot command $a_{\tau}$ by $\mathcal{D}_{a}$, i.e., $a_{\tau}=\mathcal{D}_{a}(y_{\tau})$, where $a_{\tau}$ represents the executable robot action, such as translation, rotation, and gripper control.

\begin{figure}[t]
    \centering

   \begin{subfigure}{0.9\linewidth}
        \centering
        \includegraphics[width=\linewidth]{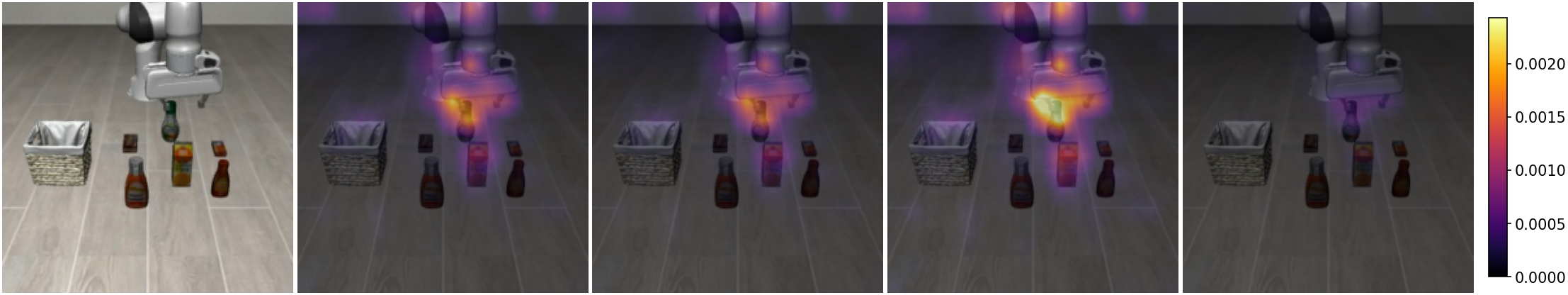}
        \hspace*{-2.5em}
        \begin{tabular*}{0.86\linewidth}{@{\extracolsep{\fill}}ccccc}
        \scriptsize \textbf{Input ($\tau=28$)} &
        \scriptsize FP: 0.000 &
        \scriptsize W8: 0.138 &
        \scriptsize W4: 0.372 &
        \scriptsize W2: 1.372
        \end{tabular*}
    \end{subfigure}

    \begin{subfigure}{0.9\linewidth}
        \centering
        \includegraphics[width=\linewidth]{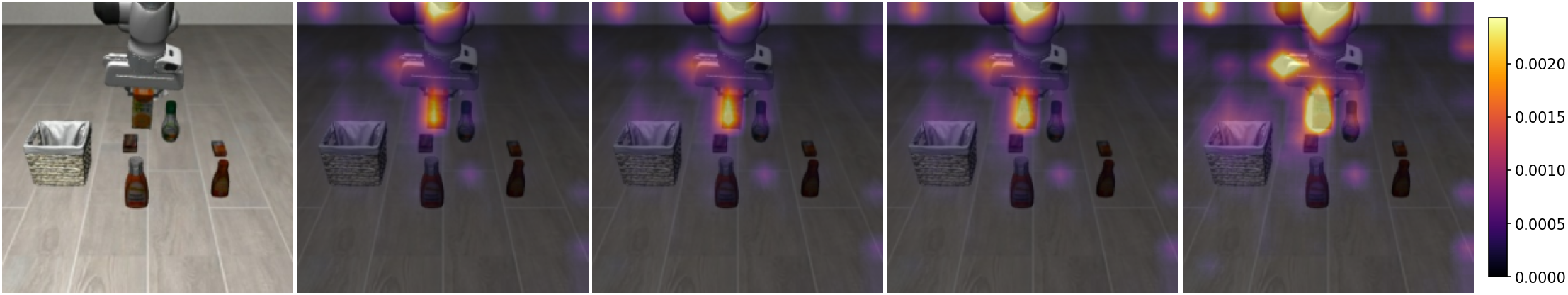}
        \hspace*{-2.5em}
        \begin{tabular*}{0.86\linewidth}{@{\extracolsep{\fill}}ccccc}
        \scriptsize \textbf{Input ($\tau=85$)} &
        \scriptsize FP: 0.000 &
        \scriptsize W8: 0.156 &
        \scriptsize W4: 0.265 &
        \scriptsize W2: 0.859
        \end{tabular*}
    \end{subfigure}

       \caption{
    Visualization of gradient-weighted task-evidence distributions at the vision-encoder output boundary when 
    \texttt{vision\_backbone.featurizer.blocks.6.mlp.fc2} is quantized. 
    The RGB observation and language prompt, ``pick up the orange juice and place it in the basket,'' are shared across FP and quantized settings; only the task-evidence overlay changes. 
    Heatmaps show the spatial allocation of internal task evidence, and the values below each image denote the corresponding task-evidence loss $\ell^{\mathrm{ev}}$. 
    Higher $\ell^{\mathrm{ev}}$ indicates greater distortion from the full-precision evidence distribution.
}
    \label{fig:task_evidence_heatmap_t28_t85}
\end{figure}

\subsection{Task-Evidence Layer Sensitivity}

Let $\theta_{\mathrm{FP}}$ denote the full-precision VLA model, and let $\theta_{m,b}$ denote the variant obtained by quantizing layer $m$ to bit-width $b$ while keeping all other layers fixed. For each candidate pair $(m,b)$, Mix-QVLA computes a task-evidence sensitivity score by comparing the evidence produced by $\theta_{\mathrm{FP}}$ and $\theta_{m,b}$ with respect to the same full-precision action-token sequence across selected VLA functional boundaries. The score is computed in three steps: fixing the full-precision action-token sequence as the reference decision, constructing gradient-weighted task-evidence maps at each boundary, and aggregating full-precision versus quantized evidence distortions into a layer-wise sensitivity score.

\paragraph{Reference action-support objective.}
For each calibration sample $i$, we consider a fixed VLA state selected from an environment timestep $\tau_i$. We write this state as 
$z_i \equiv z_{\tau_i}=(V_{\tau_i},x_{\tau_i},P_i)$, where $V_{\tau_i}$ is the visual observation, $x_{\tau_i}$ is the robot state, and $P_i$ is the task instruction. The full-precision model defines a reference action-token sequence for this state:
\[
    y_i^\ast
    \equiv
    y_{\tau_i}^{\mathrm{FP}}
    =
    (y_{i,1}^{\ast},\ldots,y_{i,K}^{\ast}).
\]
This sequence is computed once using $\theta_{\mathrm{FP}}$ and then kept fixed when evaluating all quantized variants. Using this fixed reference, we measure how strongly an evaluated model
$\theta \in \{\theta_{\mathrm{FP}},\theta_{m,b}\}$
supports the full-precision decision through the teacher-forced log-probability objective
\begin{equation}
    J_i(\theta;y_i^\ast)
    =
    \frac{1}{K}
    \sum_{k=1}^{K}
    \log p_{\theta}
    \left(
        y_{i,k}^{\ast}
        \mid
        y_{i,<k}^{\ast}; z_i
    \right),
    \label{eq:reference_action_support}
\end{equation}
where $y_{i,<k}^{\ast}=(y_{i,1}^{\ast},\ldots,y_{i,k-1}^{\ast})$ is the full-precision action-token prefix. 

\paragraph{Boundary-local task evidence.}
We evaluate task evidence at four functional boundaries, $\Gamma=\{\nu,\beta,\psi,\alpha\}$,
corresponding to the vision-encoder output, projector output, language-policy representation, and action head representation before the action-token logits, respectively. For calibration sample $i$ and evaluated model $\theta$, let $H_{i,\gamma}^{\theta}$ denote the hidden representation at boundary $\gamma\in\Gamma$.

Because different VLA boundaries have different token structures, feature dimensions, and activation scales, directly comparing raw hidden states can confound task evidence with representation-specific scale effects. We therefore compute evidence after boundary-local normalization using full-precision calibration statistics. For each boundary $\gamma$, let $\mu_{\gamma}^{\mathrm{FP}}$ and $\sigma_{\gamma}^{\mathrm{FP}}$ denote the mean and standard deviation estimated from full-precision calibration activations at that boundary. The normalized boundary representation is
\begin{equation}
    Z_{i,\gamma}^{\theta}
    =
    \frac{
        H_{i,\gamma}^{\theta}
        -
        \mu_{\gamma}^{\mathrm{FP}}
    }{
        \sigma_{\gamma}^{\mathrm{FP}}+\epsilon
    } .
    \label{eq:boundary_normalization}
\end{equation}

We define task evidence as the boundary features that are both active in the calibrated representation and influential for supporting the fixed full-precision action-token decision. This is measured with a gradient-weighted evidence map:
\begin{equation}
    E_{i,\gamma}^{\theta}
    =
    \left|
    Z_{i,\gamma}^{\theta}
    \odot
    \nabla_{Z_{i,\gamma}^{\theta}}
    J_i(\theta;y_i^\ast)
    \right| ,
    \label{eq:task_evidence_map}
\end{equation}
where $\odot$ denotes element-wise multiplication. The activation term identifies features present in the boundary representation, while the gradient term measures how changes in those features affect support for the reference full-precision decision.

\begin{figure}[t]
    \centering
    
    \begin{subfigure}[t]{0.33\linewidth}
        \centering
        \includegraphics[width=\linewidth]{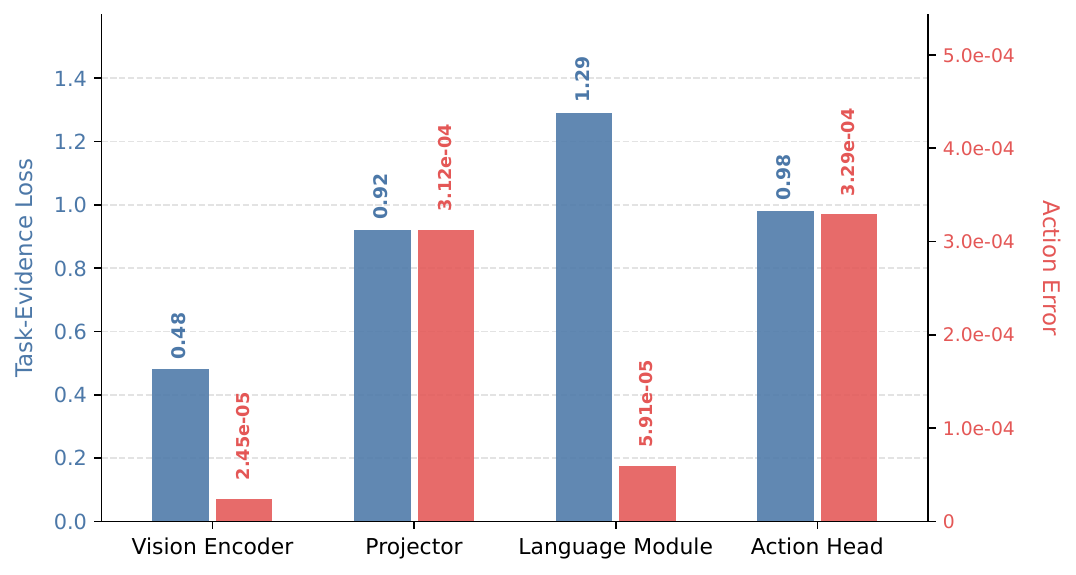}
        \caption{}
        \label{fig:boundary_vs_action}
    \end{subfigure}
    \hfill
    \begin{subfigure}[t]{0.65\linewidth}
        \centering
        \includegraphics[width=\linewidth]{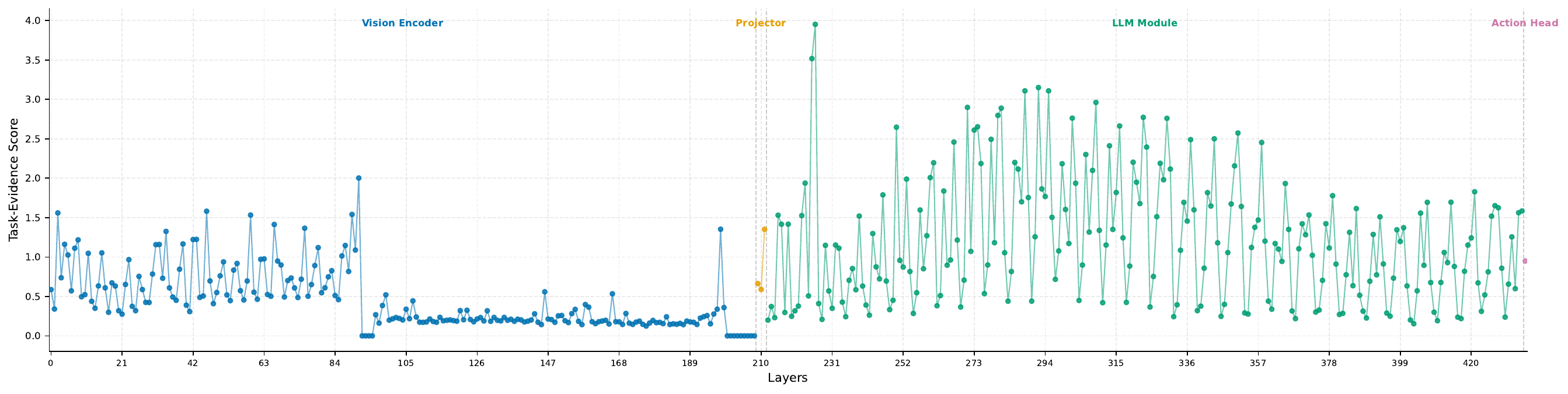}
        \caption{}
        \label{fig:global_layer_sensitivity}
    \end{subfigure}
    
    \caption{
    OpenVLA sensitivity analysis.
    \textbf{(a)} Comparison between task-evidence and action error under module-wise VLA quantization.
    They produce different boundary rankings: the language module exhibits the largest task-evidence loss despite relatively small action error, showing that action-only sensitivity can miss internal evidence degradation.
    \textbf{(b)} Global layer-wise task-evidence sensitivity across the vision encoder, projector, LLM module, and action head.
}
    \label{fig:boundary_and_global_sensitivity}
\end{figure}

\paragraph{Boundary evidence consistency.}
Given the full-precision evidence map $E_{i,\gamma}^{\theta_{\mathrm{FP}}}$ and the quantized evidence map $E_{i,\gamma}^{\theta_{m,b}}$, we measure how much quantization distorts the evidence at boundary $\gamma$. We decompose this distortion into two complementary aspects: evidence mass and evidence attribution. Evidence mass captures the overall strength of decision-supporting evidence, while evidence attribution captures how that evidence is distributed across tokens, channels, or features within the same boundary. First, we define the evidence mass at boundary $\gamma$ as
\begin{equation}
    M_{i,\gamma}^{\theta}
    =
    \frac{1}{d_{\gamma}}
    \sum_{j=1}^{d_{\gamma}}
    E_{i,\gamma,j}^{\theta},
    \label{eq:evidence_mass}
\end{equation}
where $d_{\gamma}$ is the number of elements in the boundary representation. The evidence-mass distortion caused by quantizing layer $m$ to bit-width $b$ is
\begin{equation}
    \Delta_{i,\gamma}^{\mathrm{mass}}(m,b)
    =
    \left|
    \log
    \frac{
        M_{i,\gamma}^{\theta_{m,b}}+\epsilon
    }{
        M_{i,\gamma}^{\theta_{\mathrm{FP}}}+\epsilon
    }
    \right|.
    \label{eq:evidence_mass_distortion}
\end{equation}
This term penalizes changes in the total amount of evidence supporting the full-precision decision. Second, we compare the internal allocation of evidence within the boundary by normalizing each evidence map into an attribution profile and measuring its Jensen--Shannon divergence:
\begin{equation}
\begin{aligned}
    a_{i,\gamma,j}^{\theta}
    &=
    \frac{
        E_{i,\gamma,j}^{\theta}+\epsilon
    }{
        \sum_{j'=1}^{d_{\gamma}}
        \left(E_{i,\gamma,j'}^{\theta}+\epsilon\right)
    }, \\
    \Delta_{i,\gamma}^{\mathrm{attr}}(m,b)
    &=
    D_{\mathrm{JS}}
    \left(
        a_{i,\gamma}^{\theta_{\mathrm{FP}}},
        a_{i,\gamma}^{\theta_{m,b}}
    \right).
\end{aligned}
\label{eq:evidence_attr_profile_and_distortion}
\end{equation}
This term captures whether quantization reallocates the decision-supporting evidence to different tokens, channels, or features, even when the total evidence mass remains similar. We combine the two distortions into a boundary-level task-evidence loss:
\begin{equation}
    \ell_{i,\gamma}^{\mathrm{ev}}(m,b)
    =
    \Delta_{i,\gamma}^{\mathrm{mass}}(m,b)
    +
    \lambda
    \Delta_{i,\gamma}^{\mathrm{attr}}(m,b),
    \label{eq:boundary_evidence_loss}
\end{equation}
where we set $\lambda=1$ to give equal weight to evidence-mass and attribution-distribution distortion. A larger $\ell_{i,\gamma}^{\mathrm{ev}}(m,b)$ indicates stronger degradation of the full-precision evidence pathway at boundary $\gamma$.

\paragraph{Layer sensitivity.}
The VLA decision pathway is structured across multiple functional stages, so boundary-local evidence losses must be aggregated into a single layer sensitivity score. A simple average can dilute a localized failure at a critical boundary. Multiplicative retention can emphasize pathway survival, but may saturate when several boundaries are moderately degraded. Mix-QVLA therefore uses a soft-bottleneck aggregation. This objective allows a strongly degraded boundary to have high influence on the final score while still accounting for moderate degradation across other boundaries:
\begin{equation}
    L_{i}^{\mathrm{SB}}(m,b;\kappa)
    =
    \kappa
    \log
    \left(
    \frac{1}{|\Gamma_i|}
    \sum_{\gamma \in \Gamma_i}
    \exp
    \left(
    \frac{
    \ell_{i,\gamma}^{\mathrm{ev}}(m,b)
    }{\kappa}
    \right)
    \right),
    \label{eq:soft_bottleneck}
\end{equation}
where $\Gamma_i$ denotes the set of valid boundaries for sample $i$, and $\kappa$ is the soft-bottleneck temperature. 
We set $\kappa=0.1$ in all experiments, which keeps the aggregation close to a boundary-level bottleneck while remaining numerically smooth.
As $\kappa \to 0$, $L_i^{\mathrm{SB}}$ approaches the maximum boundary loss; for larger $\kappa$, the aggregation becomes smoother and more average-like across boundaries.

The final task-evidence sensitivity score for layer $m$ at bit-width $b$ is obtained by averaging over the fixed calibration set $\mathcal{C}=\{z_i\}_{i=1}^{N}$:
\begin{equation}
    \Omega(m,b;\kappa)
    =
    \frac{1}{N}
    \sum_{i=1}^{N}
    L_i^{\mathrm{SB}}(m,b;\kappa).
    \label{eq:layer_sensitivity_score}
\end{equation}
A higher $\Omega(m,b;\kappa)$ indicates that quantizing layer $m$ to bit-width $b$ causes stronger degradation of the task evidence supporting the full-precision policy decision. 
This score is used in the subsequent mixed-precision allocation stage to prioritize higher precision for layers that are most critical to preserving the full-precision decision pathway

\begin{figure*}[t]
    \centering

    \begin{subfigure}[t]{0.48\textwidth}
        \centering
        \includegraphics[width=\linewidth]{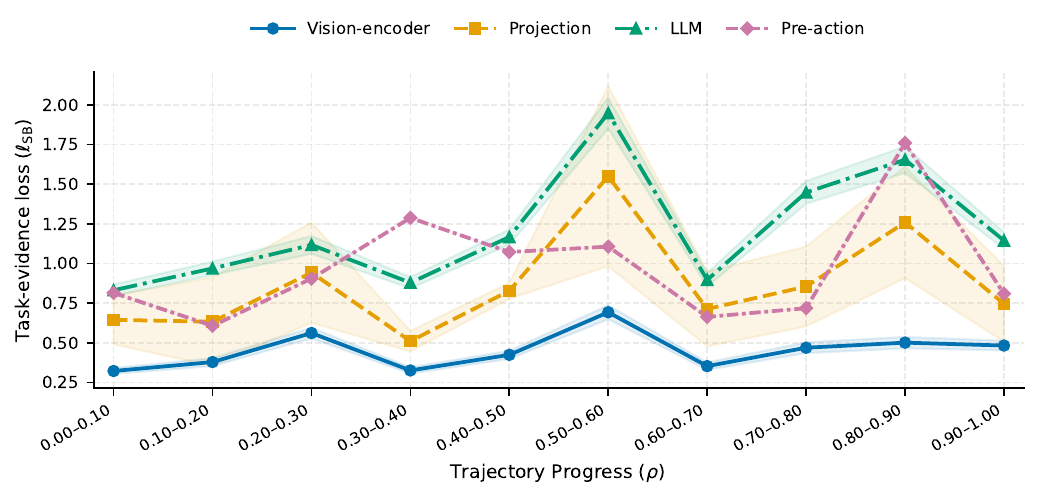}
        \caption{Boundary-wise temporal sensitivity.}
        \label{fig:boundary_temporal_w4}
    \end{subfigure}
    \hfill
    \begin{subfigure}[t]{0.48\textwidth}
        \centering
        \includegraphics[width=\linewidth]{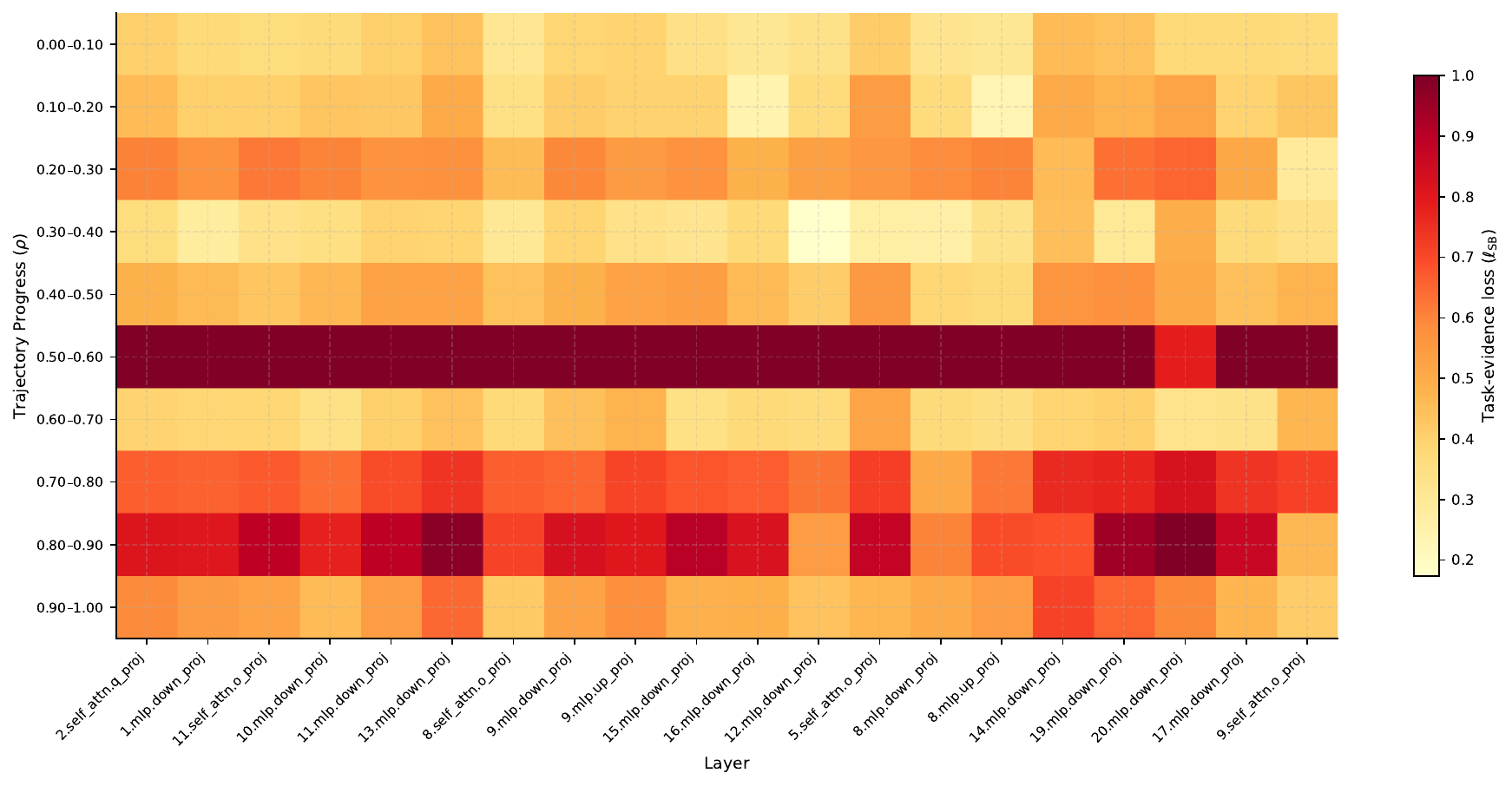}
        \caption{Top-20 sensitive layers.}
        \label{fig:topk_temporal_w4}
    \end{subfigure}

        \caption{
    Temporal task-evidence sensitivity under layer-wise 4-bit quantization.
    For each candidate layer, only that layer is quantized, while the rest of the VLA model remains full precision.
    (a) Boundary-wise task-evidence loss over normalized trajectory progress, with shaded regions indicating temporal variability.
    (b) Temporal sensitivity of the top-20 sensitive layers.
}
    \label{fig:temporal_sensitivity_w4}
\end{figure*}

\subsection{Temporal Task-Evidence Layer Sensitivity}

The task-evidence sensitivity score in Eq.~\eqref{eq:layer_sensitivity_score} provides a global estimate of quantization fragility, but it does not indicate when the degradation occurs within a manipulation trajectory. This temporal information is important for VLA policies, since early steps may rely more on visual grounding and object localization, whereas later steps may depend more on action formation and fine control. We therefore retain the trajectory index and timestep of each calibration state and use the per-state soft-bottleneck loss \(L_i^{\mathrm{SB}}(m,b;\kappa)\) as a temporal sensitivity signal. Let calibration sample \(i\) come from trajectory \(r_i\) at timestep \(\tau_i\), and let \(T_{r_i}\) denote the length of that trajectory. We map each sample to a normalized trajectory progress value
\begin{equation}
    \rho_i
    =
    \frac{\tau_i}{T_{r_i}-1}
    \in [0,1].
\end{equation}

We divide \([0,1]\) into \(Q\) temporal bins and let \(\mathcal{C}_q\) denote the calibration samples whose normalized progress falls into bin \(q\). For each non-empty bin, the phase-wise task-evidence sensitivity is
\begin{equation}
    \Omega_q(m,b;\kappa)
    =
    \frac{1}{|\mathcal{C}_q|}
    \sum_{i\in \mathcal{C}_q}
    L_i^{\mathrm{SB}}(m,b;\kappa).
\end{equation}
This score measures how strongly quantizing layer \(m\) to bit-width \(b\) disrupts the full-precision evidence pathway during a specific stage of the task. To obtain a scalar temporal sensitivity for bit allocation, we use the maximum phase-wise degradation:
\begin{equation}
    \Omega_{\tau}(m,b;\kappa)
    =
    \max_{q:|\mathcal{C}_q|>0}
    \Omega_q(m,b;\kappa).
\end{equation}
This term captures temporally localized evidence disruption and is computed only during offline calibration.

\subsection{Mixed-Precision Bit Allocation}

Given the task-evidence sensitivity \(\Omega(m,b;\kappa)\) and temporal task-evidence sensitivity \(\Omega_{\tau}(m,b;\kappa)\), Mix-QVLA assigns one bit-width to each quantizable layer. Let \(\mathcal{M}\) denote the set of quantizable layers and \(\mathcal{B}=\{2,4,8,16\}\) denote the candidate bit-widths. We introduce a binary assignment variable \(x_{m,b}\in\{0,1\}\), where \(x_{m,b}=1\) indicates that layer \(m\) is assigned bit-width \(b\). The mixed-precision bit allocation is formulated as:
\begin{subequations}
\label{eq:mp_allocation_full}
\begin{align}
    \min_{\{x_{m,b}\}} \quad
    &
    \sum_{m\in\mathcal{M}}
    \sum_{b\in\mathcal{B}}
    x_{m,b}
    \left[
    \alpha \Omega(m,b;\kappa)
    +
    \beta \Omega_{\tau}(m,b;\kappa)
    \right]
    \label{eq:mp_allocation_obj}
    \\
    \text{s.t.} \quad
    &
    \sum_{b\in\mathcal{B}} x_{m,b}=1,
    \quad
    \forall m\in\mathcal{M},
    \label{eq:mp_onehot}
    \\
    &
    \sum_{m\in\mathcal{M}}
    \sum_{b\in\mathcal{B}}
    x_{m,b} C_{\mathrm{size}}(m,b)
    \leq
    C_{\mathrm{size}}^{\mathrm{target}},
    \label{eq:mp_size_constraint}
    \\
    &
    \sum_{m\in\mathcal{M}}
    \sum_{b\in\mathcal{B}}
    x_{m,b} C_{\mathrm{bitops}}(m,b)
    \leq
    C_{\mathrm{bitops}}^{\mathrm{target}},
    \label{eq:mp_bitops_constraint}
    \\
    &
    x_{m,b}\in\{0,1\},
    \quad
    \forall m\in\mathcal{M},\; b\in\mathcal{B}.
    \label{eq:mp_binary}
\end{align}
\end{subequations}

In Eq.~\eqref{eq:mp_allocation_obj}, the objective minimizes the combined task-evidence and temporal task-evidence degradation, where \(\alpha\) and \(\beta\) control their relative contributions. Eq.~\eqref{eq:mp_onehot} enforces exactly one bit-width per layer, while Eqs.~\eqref{eq:mp_size_constraint} and~\eqref{eq:mp_bitops_constraint} impose separate model-size and BitOps budgets. We compute \(C_{\mathrm{size}}(m,b)=N_m b\) and \(C_{\mathrm{bitops}}(m,b)=\mathrm{MACs}(m)b^2\), where \(N_m\) is the number of parameters in layer \(m\). The BitOps cost assumes W\(b\)A\(b\) quantization.

\begin{table*}[t]
\caption{
Performance comparison of quantization methods on OpenVLA and OpenVLA-OFT under different weight--activation quantization settings. 
W4A4 and W8A8 denote quantization of both weights (W) and activations (A) to 4 and 8 bits, respectively. 
Bold indicates the best performance.
}
\centering
\resizebox{\textwidth}{!}{
\begin{tabular}{ll l c c c c c c c c}
\toprule
\textbf{Model} & \textbf{Setting} & \textbf{Method} &
\textbf{Spatial} & \textbf{Object} & \textbf{Goal} & \textbf{Long} &
\textbf{Avg}$\uparrow$ & $\Delta$ & \textbf{Mem. (GB)}$\downarrow$ & \textbf{Speedup}$\uparrow$ \\
\midrule

\multirow{10}{*}{OpenVLA}
& FP Model & - & 84.7\% & 88.4\% & 79.2\% & 53.7\% & 76.5\% & - & 15.2 & 1$\times$ \\
\cmidrule(lr){2-11}
& \multirow{3}{*}{W8A8}
& SmoothQuant & 84.2\% & 87.8\% & 77.8\% & 53.2\% & 75.8\% & {-0.7\%} & 7.4 & 1.40$\times$ \\
& & OmniQuant & 82.6\% & 86.2\% & 74.8\% & 51.7\% & 73.8\% & {-2.7\%} & 7.8 & 1.26$\times$ \\
& & QVLA & \textbf{85.2}\% & 88.0\% & 77.6\% & \textbf{54.2}\% & 76.3\% & {\textbf{-0.2}\%} & 7.1 & 1.42$\times$ \\
& & \textbf{Mix-QVLA} & 84.7\% & \textbf{88.1}\% & \textbf{78.9}\% & 53.4\% & \textbf{76.3}\% & {\textbf{-0.2}\%}& \textbf{6.6} & \textbf{1.46$\times$} \\
\cmidrule(lr){2-11}
& \multirow{3}{*}{W4A4}
& SmoothQuant & 69.2\% & 73.2\% & 69.6\% & 40.9\% & 63.2\% & {-13.3\%} & 4.7 & 1.52$\times$ \\
& & OmniQuant & 82.2\% & 85.4\% & 75.4\% & 50.3\% & 73.3\% & {-3.2\%} & 5.4 & 1.43$\times$ \\
& & QVLA & 84.4\% & 87.6\% & 78.8\% & 53.0\% & 76.0\% & {-0.5\%} & 4.3 & 1.47$\times$ \\
& & DyQ-VLA & \textbf{84.7}\% &87.8\% & 78.5\% & 53.4\% & 76.1\% & {-0.4\%} & 4.7 & 1.51$\times$ \\
& & \textbf{Mix-QVLA} & \textbf{84.7}\% & \textbf{87.9}\% & \textbf{78.9}\% & \textbf{53.5}\% & \textbf{76.3}\% & {\textbf{-0.2}\%} & \textbf{4.0} & \textbf{1.52$\times$} \\
\midrule

\multirow{9}{*}{OpenVLA-OFT}
& FP Model & - & 97.6\% & 98.4\% & 97.9\% & 94.5\% & 97.1\% & - & 15.4 & 1$\times$ \\
\cmidrule(lr){2-11}
& \multirow{3}{*}{W8A8}
& SmoothQuant & 96.4\% & 97.8\% & 95.4\% & 94.3\% & 96.0\% & {-1.1\%} & 7.7 & 1.41$\times$ \\
& & OmniQuant & 95.4\% & 96.2\% & 93.0\% & 92.6\% & 94.3\% & {-2.8\%} & 8.0 & 1.30$\times$ \\
& & QVLA & 97.2\% & \textbf{98.2}\% & 95.8\% & \textbf{94.3}\% & 96.4\% & {-0.7\%} & 7.2 & 1.36$\times$ \\
& & \textbf{Mix-QVLA} & \textbf{97.4}\% & \textbf{98.2}\% & \textbf{96.4}\% & \textbf{94.3}\% & \textbf{96.6}\% & {\textbf{-0.5}\%} & \textbf{6.7} & \textbf{1.39$\times$} \\
\cmidrule(lr){2-11}
& \multirow{3}{*}{W4A4}
& SmoothQuant & 77.2\% & 70.0\% & 77.8\% & 68.6\% & 73.4\% & {-23.7\%} & 4.9 & 1.53$\times$ \\
& & OmniQuant & 95.0\% & 94.4\% & 94.0\% & 92.0\% & 93.9\% & {-3.2\%} & 5.7 & 1.37$\times$ \\
& & QVLA & 96.2\% & 97.6\% & \textbf{96.4}\% & 93.8\% & 96.0\% & {-1.1\%} & 4.5 & 1.49$\times$ \\

& & \textbf{Mix-QVLA} & \textbf{96.8}\% & \textbf{97.8}\% & \textbf{96.4}\% & \textbf{94.0}\% & \textbf{96.3}\% & {\textbf{-0.8}\%} & \textbf{4.1} & \textbf{1.52$\times$} \\

\bottomrule
\end{tabular}
}
\label{tab:qvla_results}
\end{table*}

We solve Eq.~\eqref{eq:mp_allocation_full} using CVXPY (~\cite{diamond2016cvxpy}) with the ECOS\_BB branch-and-bound solver. Since all sensitivity and cost terms are precomputed from calibration statistics, the optimization reduces to a binary linear program over \(x_{m,b}\). We pre-screen the target model-size and BitOps budgets using the feasible cost range induced by \(\mathcal{B}\) to avoid infeasible allocation settings. After optimization, the final bit-width for each layer is recovered from the active assignment:
\begin{equation}
\label{eq:mp_final_assignment}
    b_m^{\ast}
    =
    \sum_{b\in\mathcal{B}} b\,x_{m,b}^{\ast},
    \quad
    \mathcal{A}^{\ast}
    =
    \{(m,b_m^{\ast})\mid m\in\mathcal{M}\}.
\end{equation}
The final allocation \(\mathcal{A}^{\ast}\) is fixed after calibration and used for all inference timesteps; Mix-QVLA does not perform timestep-wise bit switching during deployment.

\section{Experiments}

We evaluate Mix-QVLA on LIBERO (~\cite{liu2023libero}), a standard benchmark for language-conditioned robotic manipulation. Following prior VLA evaluation protocols, we consider the four LIBERO task suites: Spatial, Object, Goal, and Long. The full-precision baseline is an OpenVLA-style policy with BF16 weights. All experiments are conducted on a single NVIDIA A100 GPU. We follow a post-training quantization setting, where the pretrained policy remains frozen. Calibration samples are drawn from LIBERO training demonstrations and include RGB observations, robot states, task instructions, and trajectory timestep indices. Mix-QVLA considers bit-widths from \(\{2,4,8,16\}\) and assigns mixed precision over all quantizable layers using the proposed task-evidence sensitivity criterion. We compare against state-of-the-art mixed-precision VLA quantization methods under matched calibration data, evaluation protocol, and bit-budget constraints. We report LIBERO success rate as the primary metric and summarize efficiency using average bit-width and compression ratio.

\subsection{Results}

\paragraph{Results on weight-activation quantization.} Table~\ref{tab:qvla_results} compares Mix-QVLA with existing quantization methods under W8A8 and W4A4 settings on OpenVLA and OpenVLA-OFT. Across both models, Mix-QVLA consistently achieves the best or competitive LIBERO average success rate while using lower GPU memory than prior methods. On OpenVLA, Mix-QVLA matches the best W8A8 average performance of \(76.3\%\) while reducing memory to \(6.6\) GB and achieving the highest speedup of \(1.46\times\). Under the more aggressive W4A4 setting, Mix-QVLA improves the average success rate to \(76.3\%\), outperforming QVLA and DyQ-VLA while reducing memory to \(4.0\) GB and achieving \(1.52\times\) speedup. Similar trends are observed on OpenVLA-OFT, where Mix-QVLA obtains the highest average success rate under both W8A8 and W4A4 settings, with \(96.6\%\) and \(96.3\%\), respectively.

\paragraph{Results on weight-only quantization.} Table~\ref{tab:weight_only_quant} reports the performance of weight-only quantization on OpenVLA and OpenVLA-OFT. Compared with AWQ, both QVLA and Mix-QVLA substantially preserve task success under W8A16 and W4A16 settings, showing that sensitivity-aware allocation is important for VLA quantization. On OpenVLA, Mix-QVLA achieves the best average performance under both W8A16 and W4A16, reaching \(76.9\%\) and \(76.6\%\), respectively, while also reducing memory to \(6.7\) GB and \(4.1\) GB. This improves over QVLA in average success rate and memory usage, especially in the W4A16 setting where Mix-QVLA maintains performance slightly above the BF16 baseline while using substantially lower memory. Similar trends are observed on OpenVLA-OFT: Mix-QVLA matches or improves task success across the four LIBERO suites and achieves the best average performance under both W8A16 and W4A16, with lower memory than QVLA.

\begin{table*}[t]
\centering
\caption{
Performance comparison of quantization methods on OpenVLA and OpenVLA-OFT under weight-only quantization settings.
Weight-only quantization mainly reduces model memory footprint and typically provides limited latency improvement.
Bold values indicate the best performance.
}
\label{tab:weight_only_quant}
\resizebox{\textwidth}{!}{
\begin{tabular}{lllccccccc}
\toprule
\textbf{Model} & \textbf{Setting} & \textbf{Method} &
\textbf{Spatial} & \textbf{Object} & \textbf{Goal} & \textbf{Long} &
\textbf{Avg.} $\uparrow$ & $\Delta$ & \textbf{Mem. (GB)} $\downarrow$ \\
\midrule

\multirow{7}{*}{OpenVLA}
& FP Model & -- 
& 84.7\% & 88.4\% & 79.2\% & 53.7\% & 76.5\% & -- & 15.2 \\
\cmidrule(l){2-10}

& \multirow{3}{*}{W8A16}
& AWQ 
& 82.3\% & 87.2\% & 74.6\% & 50.7\% & 73.7\% & -1.8\% & 7.6 \\

& 
& QVLA
& \textbf{86.2}\% & 88.4\% & 79.4\% & 53.1\% & 76.8\% & +0.3\% & 7.2 \\

& 
& \textbf{Mix-QVLA}
& 86.0\% & \textbf{88.5}\% & \textbf{79.6}\% & \textbf{53.5}\% & \textbf{76.9}\% & \textbf{+0.4}\% & \textbf{6.7} \\
\cmidrule(l){2-10}

& \multirow{3}{*}{W4A16}
& AWQ 
& 80.0\% & 81.2\% & 74.6\% & 47.2\% & 70.8\% & -4.7\% & 5.0 \\

& 
& QVLA
& \textbf{86.0}\% & \textbf{88.6}\% & 78.4\% & 52.8\% & 76.5\% & +0.0\% & 4.3 \\

& 
& \textbf{Mix-QVLA}
& 85.8\% & 88.2\% & \textbf{78.8}\% & \textbf{53.4}\% & \textbf{76.6}\% & \textbf{+0.1}\% & \textbf{4.1} \\

\midrule

\multirow{7}{*}{OpenVLA-OFT}
& FP Model & -- 
& 97.6\% & 98.4\% & 97.9\% & 94.5\% & 97.1\% & -- & 15.4 \\
\cmidrule(l){2-10}

& \multirow{3}{*}{W8A16}
& AWQ 
& 95.2\% & 96.8\% & 95.4\% & 93.1\% & 95.1\% & -2.0\% & 8.0 \\

& 
& QVLA
& 97.4\% & \textbf{98.6}\% & 97.2\% & \textbf{94.6}\% & 97.0\% & -0.1\% & 7.4 \\
& 
& \textbf{Mix-QVLA}
& \textbf{97.5}\% & \textbf{98.6}\% & \textbf{97.7}\% & \textbf{94.6}\% & \textbf{97.1}\% & \textbf{+0.0}\% & \textbf{6.8} \\
\cmidrule(l){2-10}

& \multirow{3}{*}{W4A16}
& AWQ 
& 93.0\% & 92.4\% & 93.8\% & 90.7\% & 92.5\% & -4.5\% & 5.2 \\

& 
& QVLA
& 97.0\% & \textbf{98.4}\% & 96.8\% & \textbf{94.4}\% & 96.7\% & -0.4\% & 4.5 \\
& 
& \textbf{Mix-QVLA}
& \textbf{97.1}\% & \textbf{98.4}\% & \textbf{97.4}\% & \textbf{94.4}\% & \textbf{96.9}\% & \textbf{-0.2}\% & \textbf{4.2} \\
\bottomrule
\end{tabular}
}
\end{table*}

\begin{table}[t]
\centering
\caption{
Ablation studies for Mix-QVLA.
(a) Sensitivity signal ablation under matched mixed-precision allocation settings. 
Act., Ev., and Temp. denote action-based sensitivity, task-evidence sensitivity, and temporal task-evidence sensitivity, respectively.
Bit, Mem., and Avg. denote average bit-width, GPU memory usage in GB, and average LIBERO success rate.
(b) Temporal weighting ablation, where \(\alpha\) and \(\beta\) control the contributions of task-evidence sensitivity and temporal task-evidence sensitivity, respectively.
}
\label{tab:mixqvla_ablation}

\begin{minipage}{0.60\linewidth}
\centering
\scriptsize
\resizebox{\linewidth}{!}{
\begin{tabular}{cccccc}
\toprule
\textbf{Act.}
& \textbf{Ev.}
& \textbf{Temp.}
& \textbf{Bit} $\downarrow$
& \textbf{Mem.} $\downarrow$
& \textbf{Avg.} $\uparrow$ \\
\midrule
\multicolumn{3}{c}{\textbf{Full Precision}} 
& 16.0 & 15.2 & 76.5 \\
\midrule
\checkmark & --         & --         & 4.00 & 4.3 & 76.0 \\
--         & \checkmark & --         & 3.94 & 4.0 & 75.9 \\
--         & --         & \checkmark & 3.95 & 4.1 & 75.6 \\
--         & \checkmark & \checkmark & 3.96 & 4.0 & 76.3 \\
\bottomrule
\end{tabular}
}
\vspace{1mm}

\textbf{(a) Sensitivity signal ablation}
\end{minipage}
\hfill
\begin{minipage}{0.32\linewidth}
\centering
\scriptsize
\resizebox{\linewidth}{!}{
\begin{tabular}{ccc}
\toprule
\(\alpha\)
& \(\beta\)
& \textbf{Avg.} $\uparrow$ \\
\midrule
1.00 & 0.00 & 75.9 \\
0.75 & 0.25 & 76.3 \\
0.50 & 0.50 & 76.0 \\
0.25 & 0.75 & 75.8 \\
0.00 & 1.00 & 75.6 \\
\bottomrule
\end{tabular}
}
\vspace{1mm}

\textbf{(b) Temporal weighting ablation}
\end{minipage}

\end{table}

\subsection{Ablation Study}
Table~\ref{tab:mixqvla_ablation}(a) evaluates the contribution of different sensitivity signals for mixed-precision allocation. The BF16 model achieves the highest average LIBERO success rate of \(76.5\%\), while QVLA obtains \(76.0\%\) at 4-bit average precision and 4.3 GB memory. Using only task-evidence sensitivity gives a comparable success rate of \(75.9\%\), and using only temporal sensitivity gives \(75.6\%\), indicating that each signal provides useful but incomplete guidance. By combining task-evidence and temporal sensitivity, Mix-QVLA improves the average success rate to \(76.3\%\), while maintaining a lower average bit-width of 3.96 and 4.0 GB memory. 

Table~\ref{tab:mixqvla_ablation}(b) further studies the weighting between task-evidence sensitivity and temporal sensitivity in the bit-allocation objective. The task-evidence-only setting \((\alpha=1.0,\beta=0.0)\) achieves \(75.9\%\), while the temporal-only setting \((\alpha=0.0,\beta=1.0)\) drops to \(75.6\%\), showing that temporal information should complement rather than replace the global task-evidence signal. The best result is obtained with a moderate temporal contribution \((\alpha=0.75,\beta=0.25)\), reaching \(76.3\%\). Increasing the temporal weight further reduces performance, suggesting that over-emphasizing worst-phase degradation may over-protect temporally sensitive layers while under-protecting layers with consistent global importance. Overall, the ablation confirms that Mix-QVLA benefits from a balanced allocation signal that preserves both average task evidence and temporally localized evidence.

\section{Limitations}
Mix-QVLA is evaluated on OpenVLA-style policies and LIBERO simulation tasks, and further validation is needed on real-robot deployment and other VLA architectures. The task-evidence score requires additional calibration-time forward and backward passes, increasing offline analysis cost compared with action-only sensitivity. The current allocation is fixed after calibration and does not perform timestep-wise precision switching during deployment. Finally, the evidence maps are diagnostic signals rather than causal guarantees of task success, so they should be interpreted together with rollout performance.

\section{Conclusion}
We presented Mix-QVLA, a task-evidence-aware mixed-precision quantization framework for vision-language-action models. Instead of relying only on final action deviation, Mix-QVLA measures whether low-bit quantization preserves the internal task-evidence supporting the full-precision policy decision. By combining task-evidence sensitivity with temporal sensitivity across manipulation trajectories, the proposed allocation strategy protects task-critical layers while assigning lower precision to quantization-tolerant layers. Experiments on LIBERO with OpenVLA and OpenVLA-OFT show that Mix-QVLA maintains competitive task success under aggressive W4A4 and weight-only quantization settings, while reducing GPU memory usage compared with prior quantization baselines. 









\end{document}